\documentclass[letterpaper, 10 pt, conference]{ieeeconf}  % Comment this line out if you need a4paper

\IEEEoverridecommandlockouts                              % Needed for \thanks
\usepackage{color}
\usepackage{soul}
\usepackage{graphicx}
\usepackage{amsmath}
\usepackage{microtype}
\usepackage{float}

\title{\LARGE \bf
TaSA: Two-Phased Deep Predictive Learning of Tactile Sensory Attenuation for Improving In-Grasp Manipulation
}

\author{Pranav Ponnivalavan, Satoshi Funabashi, Alexander Schmitz, Tetsuya Ogata, Shigeki Sugano%
\thanks{This research was supported by Moonshot R\&D with a grant number of JPMJMS2031.}%
\thanks{All authors are with Waseda University, Okubo 3-4-1, Shinjuku, Tokyo 169-8555, Japan (pranavponni@fuji.waseda.jp).}%
}

\begin{document}

\maketitle
\thispagestyle{empty}
\pagestyle{empty}

%%%%%%%%%%%%%%%%%%%%%%%%%%%%%%%%%%%%%%%%%%%%%%%%%%%%%%%%%%%%%%%%%%%%%%%%%%%%%%%%
\begin{abstract}

Humans can achieve diverse in-hand manipulations, such as object pinching and tool use, which often involve simultaneous contact between the object and multiple fingers. This is still an open issue for robotic hands because such dexterous manipulation requires distinguishing between tactile sensations generated by their self-contact and those arising from external contact. Otherwise, object/robot breakage happens due to contacts/collisions. Indeed, most approaches ignore self-contact altogether, by constraining motion to avoid/ignore self-tactile information during contact. While this reduces complexity, it also limits generalization to real-world scenarios where self-contact is inevitable. Humans overcome this challenge through self-touch perception, using predictive mechanisms that anticipate the tactile consequences of their own motion, through a principle called sensory attenuation, where the nervous system differentiates predictable self-touch signals, allowing novel object stimuli to stand out as relevant. 
Deriving from this, we introduce {TaSA}, a two-phased deep predictive learning framework. In the first phase, TaSA explicitly learns self-touch dynamics, modeling how a robot’s own actions generate tactile feedback. In the second phase, this learned model is incorporated into the motion learning phase, to emphasize object contact signals during manipulation. We evaluate TaSA on a set of insertion tasks, which demand fine tactile discrimination: inserting a pencil lead into a mechanical pencil, inserting coins into a slot, and fixing a paper clip onto a sheet of paper, with various orientations, positions, and sizes. Across all tasks, policies trained with TaSA achieve significantly higher success rates than baseline methods, demonstrating that structured tactile perception with self-touch based on sensory attenuation is critical for dexterous robotic manipulation.
\end{abstract}

%%%%%%%%%%%%%%%%%%%%%%%%%%%%%%%%%%%%%%%%%%%%%%%%%%%%%%%%%%%%%%%%%%%%%%%%%%%%%%%%
\section{Introduction}

\begin{figure}[ht]
    \centering
    \includegraphics[width=0.9\columnwidth]{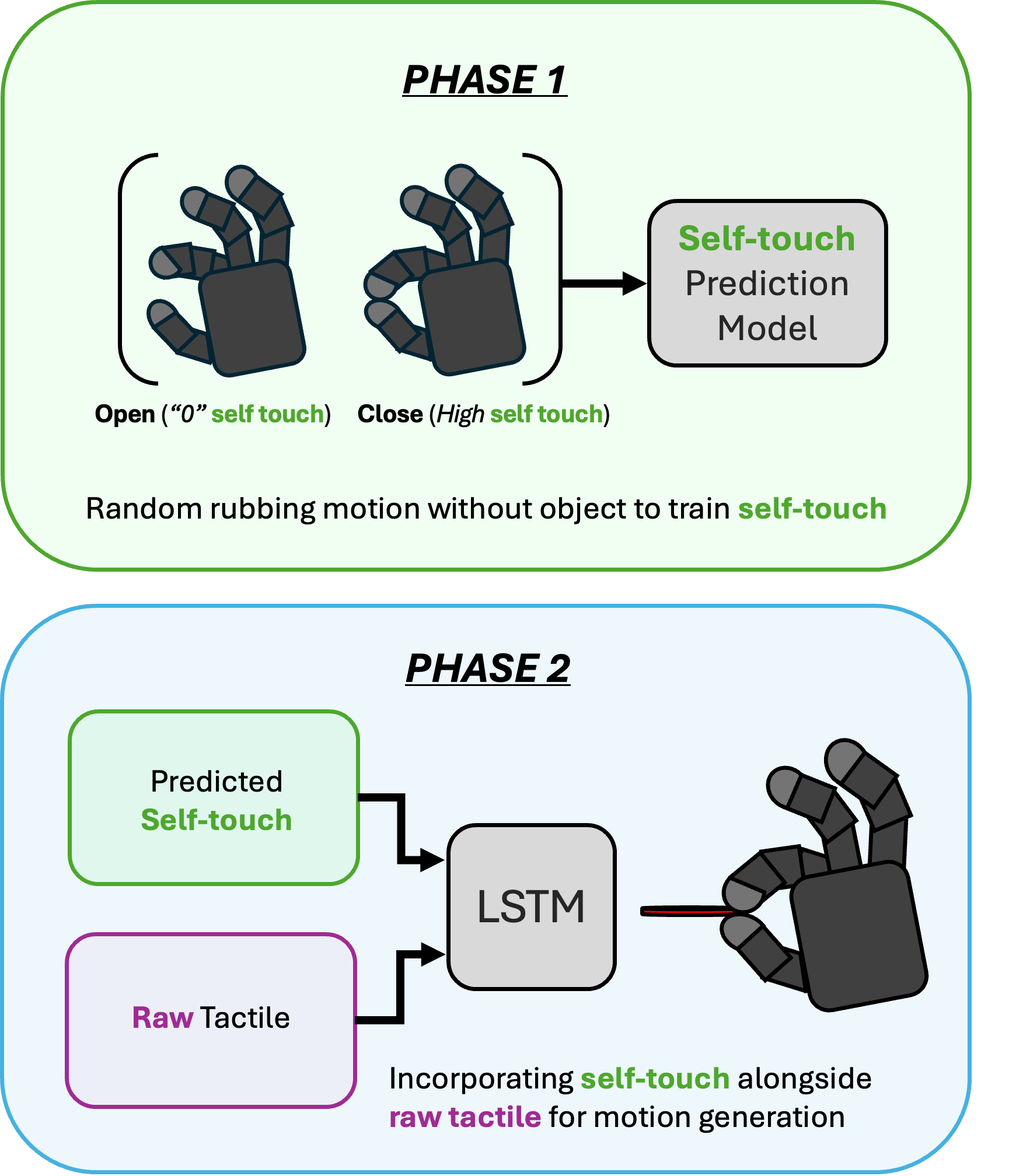}
    \caption{Diagrammatic representation of TaSA - the proposed two-phased deep predictive learning method.}
    \label{fig:motion_phase}
\end{figure}

A fundamental principle in human perception is \emph{sensory attenuation} (SA). When we perform voluntary movements, the brain predicts the sensory consequences of these actions and down-weights those that match the prediction, reducing their perceptual prominence \cite{1}. This mechanism prevents the nervous system from being overloaded by expected inputs and allows it to remain highly sensitive to unexpected or externally generated events. Experiments show that even short delays or changes in sensory feedback can disturb attenuation, highlighting how precise and adaptable it is \cite{2}. This temporal sensitivity is directly relevant for robotic manipulation, where precise timing between motor commands and tactile state is critical, because any mismatch can cause the system to misclassify self-touch as external contact, leading to unstable grasps or errors in object handling. Moreover, Kilteni and Ehrsson demonstrated that when participants experienced reduced body ownership, for example, when a seen hand was not perceived as their own, the normal attenuation of self-generated touch diminished, indicating that SA is not only a predictive motor mechanism but also critically shaped by body ownership and the subjective experience of agency \cite{3}.

For robotics, this principle motivates further investigation. Self-touch in robotic hands are when fingers contact each other or the palm. Rather than ignoring it, self-touch provides a predictable signal that can be modeled ensuring expected contacts are filtered while novel ones remain salient. By doing so, the system prevents collisions from confusing the tactile stream while simultaneously creating a clearer channel for detecting object interactions. Without understanding self-touch, robotic hands risk being overwhelmed by the generated tactile signals. Self-touch is therefore not only a by-product of multi-finger actions but also a powerful phenomenon for improving perception and control.

This becomes especially important in \emph{in-hand manipulation}, where an object is stabilized, repositioned, or reoriented within the grasp through coordinated finger motions. Prior work has demonstrated the value of tactile sensing for such tasks: Funabashi et al.\ achieved stable manipulation with CNN-LSTM models \cite{4}, but avoided finger–finger and finger–palm contacts by omitting or not realizing that potential of using it effectively. Graph convolutional networks have been applied to capture object properties from distributed tactile data \cite{5}, yet they focus on recognition rather than contact-rich control. Guzey \emph{et al.} showed that visual incentives can guide tactile dexterity \cite{6}, and later demonstrated that large-scale robotic play enables self-supervised pre-training of tactile representations \cite{7}. Both focus on transferable features rather than the sensor resolution needed to disambiguate fine contact events. Ueno et al.\ combined transformers and LSTMs for multi-finger manipulation \cite{8}, but unlike TaSA, their model lacked a mechanism to differentiate self-touch from object contact.

These works show that tactile learning enhances in-hand manipulation, but also reveal a critical gap: multi-finger manipulation inevitably produces collisions, and existing methods fail to model them explicitly. Our contribution is to fill this gap by introducing \textbf{TaSA}: a sensory attenuation-based framework that predicts and uses self-touch and we demonstrate its effectiveness across three precision insertion tasks: paper-clip fixing, coin insertion, and pencil lead insertion, comparing with raw tactile-only baseline.

\section{Related Work}

\subsection{Sensory Attenuation and Self-Touch}
The concept of sensory attenuation (SA) describes how humans down-weight predictable sensations from self-generated actions to prioritize unexpected external stimuli. This mechanism has been extended into robotics as a way to filter tactile input during self-contact. Hara \emph{et al.} demonstrated that voluntary self-touch modulates the sense of body ownership in humans, providing evidence for the perceptual consequences of attenuated self-generated signals \cite{9}. Building on this, Lanillos and Cheng proposed predictive coding architectures that allow robots to estimate and adaptively attenuate their own sensorimotor feedback \cite{10}. Nguyen \emph{et al.} further surveyed models for sensorimotor representation learning, highlighting predictive mechanisms as a key principle in active robotic systems \cite{11}. Complementing these works, Gama \emph{et al.} showed how humanoid robots can use goal-directed self-touch exploration to improve internal calibration through tactile feedback \cite{12}. In this context, calibration refers to aligning the robot’s internal kinematic and proprioceptive model with its tactile sensory readings, ensuring that the perceived contact location and force match the robot’s actual configuration. 

Although these studies establish self-touch and attenuation as useful tools for robust tactile filtering, related progress in in-hand manipulation tasks, such as insertion motions, shows the importance of further extending these ideas. Azulay \emph{et al.} applied visuotactile-based learning for socket-type insertion with compliant hands \cite{13}, but their approach relied heavily on vision, making it less robust under occlusion or uncertainty where tactile prediction is most critical. Nozu and Shimonomura demonstrated robotic bolt insertion with in-hand localization and force sensing \cite{14}, yet their strategy was largely rule-based and struggled with generalization to new geometries or tolerances. Miyama \emph{et al.} designed a five-fingered robotic hand with full tactile coverage to aid insertion motions \cite{15}; while this hardware improves contact detection, it does not incorporate predictive models that anticipate tactile dynamics.  

\subsection{Tactile Sensing for Manipulation}
High-resolution tactile sensors are essential for dexterous robotic manipulation, providing tri-axial force measurements, slip detection, and distributed contact maps. Recent advances illustrate the breadth of design approaches. Gong \emph{et al.} proposed a flexible bionic tactile sensor using solid–liquid composites to achieve three-axis measurements on deformable surfaces \cite{16}, offering adaptability but at the cost of stability and durability under repeated precision contacts. Bhirangi \emph{et al.} introduced AnySkin, a modular tactile skin system mountable across varied robot geometries \cite{17}, extending tactile coverage but with relatively low density and larger-scale patches that make it less suitable for fingertip precision tasks.
In contrast, Tomo \emph{et al.}’s uSkin fingertip sensor provides a compact, high-density array of 30 tri-axial taxels on a curved surface \cite{18}. Its millimeter-scale spacing and small form factor capture subtle force distributions at the level of finger–finger or finger–palm contacts—critical in insertion tasks where distinguishing self-touch from object contact prevents false task completion. Vision-based sensors like GelSight excel at surface geometry but miss such fine force changes due to limited sensitivity to small marker deformations \cite{19}, whereas uSkin directly measures tri-axial forces with fingertip-level precision. This makes uSkin particularly suited to our work, enabling robust modeling of self-touch while preserving reliable object contact signals for control.

\subsection{Deep Predictive Learning for Manipulation}
Alongside advances in tactile sensors, deep learning has become central to predictive control in dexterous manipulation. Sievers \emph{et al.} showed that purely tactile input with recurrent models can sustain in-hand manipulation \cite{20}, while Ichiwara \emph{et al.} combined tactile and visual cues to improve flexible handling \cite{21}. Kawaharazuka \emph{et al.} introduced parametric-bias predictors to adapt to temporal changes in contact-rich tasks \cite{22}, and Suzuki \emph{et al.} extended predictive models toward sensory attenuation by accounting for temporal delays in action–outcome coupling \cite{23}. More recent studies explicitly target multi-finger manipulation: Mack \emph{et al.} demonstrated that even low-resolution tactile input improves visuo-tactile pose estimation in multi-finger hands when vision is unreliable \cite{24}, Yin \emph{et al.} trained reinforcement learning policies for in-hand translation using tactile skin with shear and normal sensing \cite{25}, and Yin \emph{et al.} also introduced geometric retargeting for efficient human-to-robot motion transfer \cite{26}. 
Despite these advances, several limitations remain. A recurring issue is that most prior studies treat all tactile inputs as external, leaving finger–finger and finger–palm collisions unmodeled. This is problematic in contact-rich settings, where self-touch can easily mimic object contact and lead to unstable control leading to object breakage or harsh handling. Another gap is methodological: many works direct their learning objectives toward object state estimation or temporal robustness but not toward filtering self-generated tactile signals. Even advanced recurrent models with multi modal fusion lack explicit mechanisms for disambiguating self-touch from external interactions in multi-finger manipulation.

\begin{figure}[ht]
    \centering
    \includegraphics[width=0.9\columnwidth]{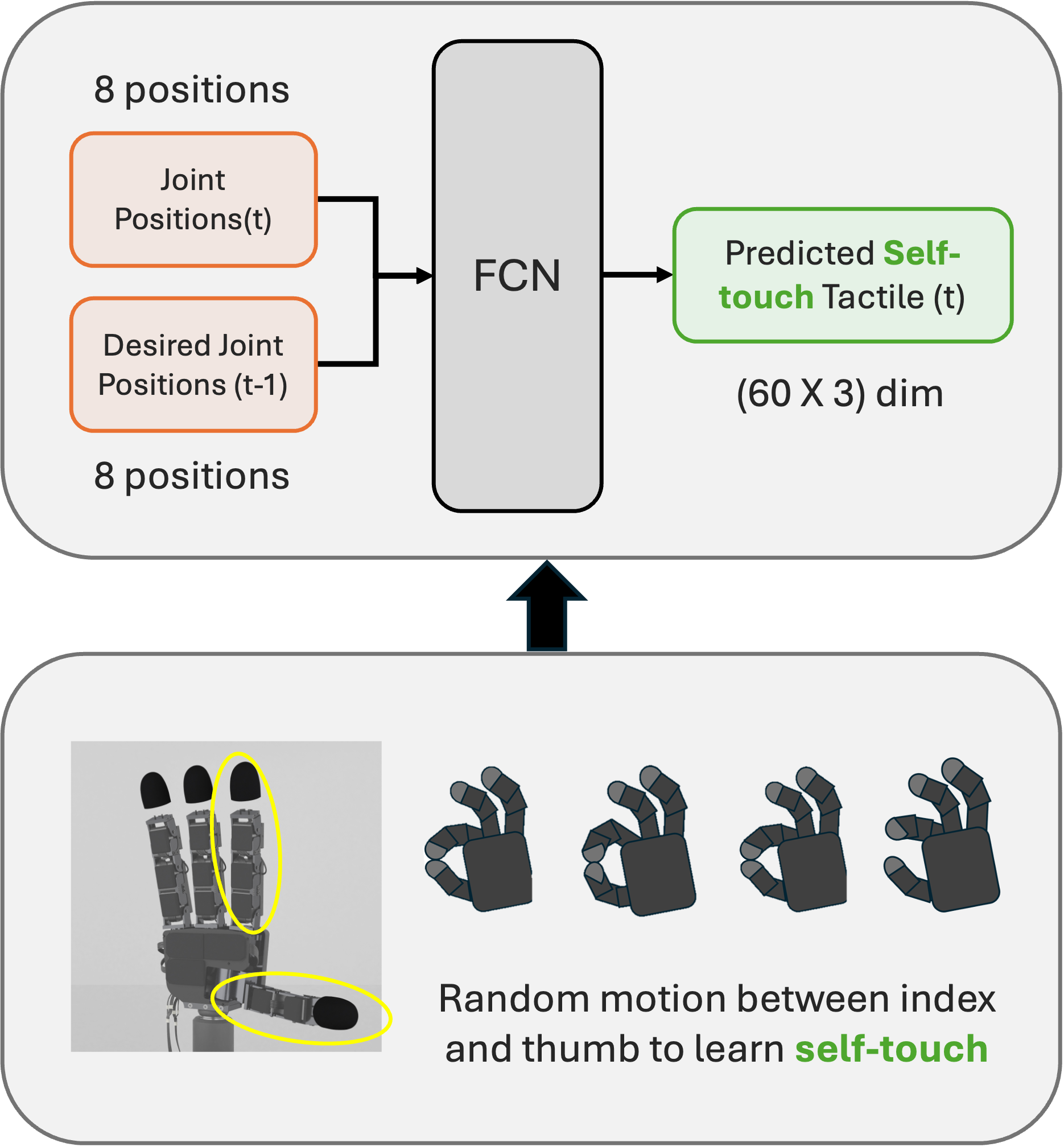}
    \caption{Diagrammatic representation of the self-touch learning phase with input joint positions and tactile data.}
    \label{fig:selftouch_phase}
\end{figure}

\section{Method}
This section presents the proposed framework and is structured into two stages: (1) a \textit{Self-Touch Learning Phase}, where a fully connected network (FCN) is trained to predict tactile feedback arising solely from finger-finger contacts, and (2) a \textit{Motion Learning Phase}, where a temporal model integrates raw tactile sensing, and predicted self-touch signals to enable robust manipulation.

\subsection{Key Idea}
The core idea of our approach is to predict self-touch in order to separate self generated tactile sensations from object-induced contact. This is achieved through two phases. In the first phase, a self-touch model is trained to map joint configurations to the resulting tactile responses from finger–finger or finger–palm contact. In the second phase, this model is reused in motion learning: using the separated self-touch, leaving a cleaner understanding of the external channel that highlights true object interactions.
Within the motion learning model, an LSTM cell processes joint states, tactile input, and predicted self-touch to capture temporal dependencies and predict the next joint configuration, motion targets, and tactile states. A frozen copy of the self-touch model provides future self-touch estimates from predicted postures, ensuring that the network reconstructs raw tactile output while maintaining a separation between external and self-induced contact. By reasoning jointly over raw tactile input, and self-touch prediction, the network develops an attenuation-based representation that supports dexterous manipulation: it can recognize when contact arises from self-touch and maintain stability, while responding differently to object-induced forces during interaction.

\begin{figure}[ht]
    \centering
    \includegraphics[width=1.0\columnwidth]{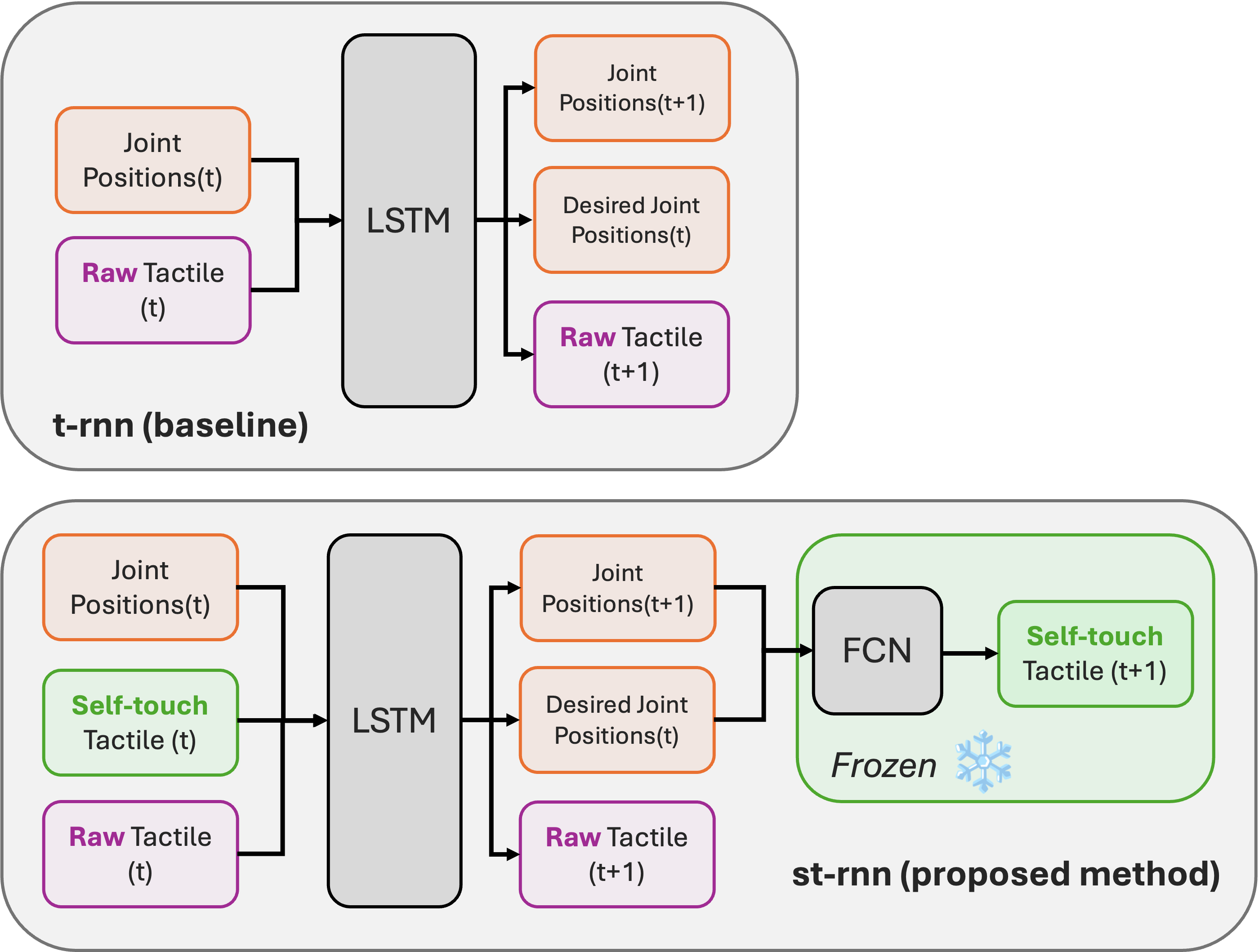}
    \caption{Comparison of the proposed method that uses both raw tactile and predicted self-touch with the baseline method that uses only raw tactile.}
    \label{fig:modelcomparison}
\end{figure}

\subsection{Self-Touch Learning Phase}
The self-touch learning phase explained in Fig.~\ref{fig:selftouch_phase} focuses on modeling the relationship between the hand’s joint configuration and its own tactile feedback, independent of external objects. At each timestep $t$, the model receives the current joint position $q_t$ (the configuration of the 8 active joints: 4 for index, 4 for thumb) and the desired joint position at the previous timestep $q_{t-1}^{des}$. These inputs are passed through a fully connected network (FCN), trained to predict the self-touch tactile state. Each fingertip sensor contains 30 taxels, and in our setup we use both the index and thumb sensors, giving a total of $30 \times 2 = 60$ taxels. Each taxel provides 3-axis force measurements $(F_x, F_y, F_z)$, resulting in tactile data of dimension $60 \times 3$. Here, hidden dimension of the FCN was set to 128 with a dropout rate of 0.2, as shown in Table. \ref{tab:selftouch_fcn}.

The FCN is optimized to minimize the prediction error between the predicted $\hat{s}_t$ and measured $s_t$ self-touch tactile signals. Since training is conducted only under free-space or self-contact motions without external objects, all tactile input during this phase corresponds to self-touch and must be predicted by the model. By training across a range of postures: from open-pinch, where fingers are separated with little or no contact, to closed-pinch and rubbing, where sustained self-contact occurs, the model learns to associate joint transitions with characteristic self-touch tactile patterns.

\begin{table}[t]
\centering
\caption{Setting of Self-Touch FCN}
\label{tab:selftouch_fcn}
\scriptsize
\setlength{\tabcolsep}{3pt}
\begin{tabular}{|p{2.5cm}|p{5.5cm}|}
\hline
\textbf{Parameter} & \textbf{Description} \\
\hline
Input & Concatenated joint positions $q_t$ and commands $q^{cmd}_t$ ($2 \times$ hand-dim = 16) \\
\hline
Hidden dim & 128 (with GELU activations and dropout 0.2) \\
\hline
Encoder & Linear(16 $\to$ 64), GELU, Dropout(0.2); Linear(64 $\to$ 128), GELU \\
\hline
Decoder & Dropout(0.2); Linear(128 $\to$ 64), GELU; Linear(64 $\to$ 188), GELU \\
\hline
Output & $\hat{s}_t^{\mathrm{idx}}$: tactile index tip (90) \\
       & $\hat{s}_t^{\mathrm{thb}}$: tactile thumb tip (90) \\
       & $\hat{q}_t$: auxiliary joint state (8) \\
\hline
\end{tabular}
\end{table}

\subsection{Motion Learning Phase}
After learning self-contact dynamics, the motion learning phase incorporates joint positions, raw tactile feedback, and predicted self-touch signals, enabling the model to generalize from finger–finger contacts to object interactions in contact-rich manipulation.

At each timestep $t$, the model receives a combined input vector
\begin{equation}
    x_t = \big[q_t, \; \hat{s}_t, \; T_t \big],
\end{equation}
where $q_t$ denotes the current joint positions, $\hat{s}_t$ the predicted self-touch tactile signal from the frozen FCN(i.e., not updated during this training phase), and $T_t$ the raw tactile sensor output. The combined input $x_t$ is processed by an LSTM cell that captures temporal dependencies across time. At each timestep, the LSTM outputs the predicted joint configuration $\hat{q}_{t+1}$, the predicted desired joint configuration for task execution $\hat{q}_t^{des}$, and the predicted raw tactile feedback $\hat{T}_{t+1}$. Encoding self-touch explicitly in the recurrent model improves the learned cause--effect relationship between motion and tactile outcomes. The LSTM’s predicted posture $\hat{q}_{t+1}$ and desired posture $\hat{q}_t^{des}$ are passed into the self-touch FCN, whose weights are frozen, to produce the next-step self-touch prediction:
\begin{equation}
    \hat{s}_{t+1} = f_{st}(\hat{q}_{t+1}, \hat{q}_t^{des}).
\end{equation}
This FCN remains fixed during training and acts as a differentiable forward model of self-contact. In addition to joint positions $q_t$, the network also receives commanded joint positions $q_t^{\mathrm{cmd}}$, which help capture the intended motion trajectory and improve prediction of the resulting self-touch signals. The output includes an auxiliary joint state $\hat{q}_t$, used as a consistency check to ensure that predicted tactile patterns remain aligned with feasible hand kinematics. Building on this foundation, the proposed method combines self-touch with raw tactile (RT) (st-rnn) and is evaluated against the baseline that uses RT only (t-rnn), as illustrated in Fig.~\ref{fig:modelcomparison} and explained in detail in Table. \ref{tab:rnn_settings}

\begin{figure}[ht]
    \centering
    \includegraphics[width=0.8\columnwidth]{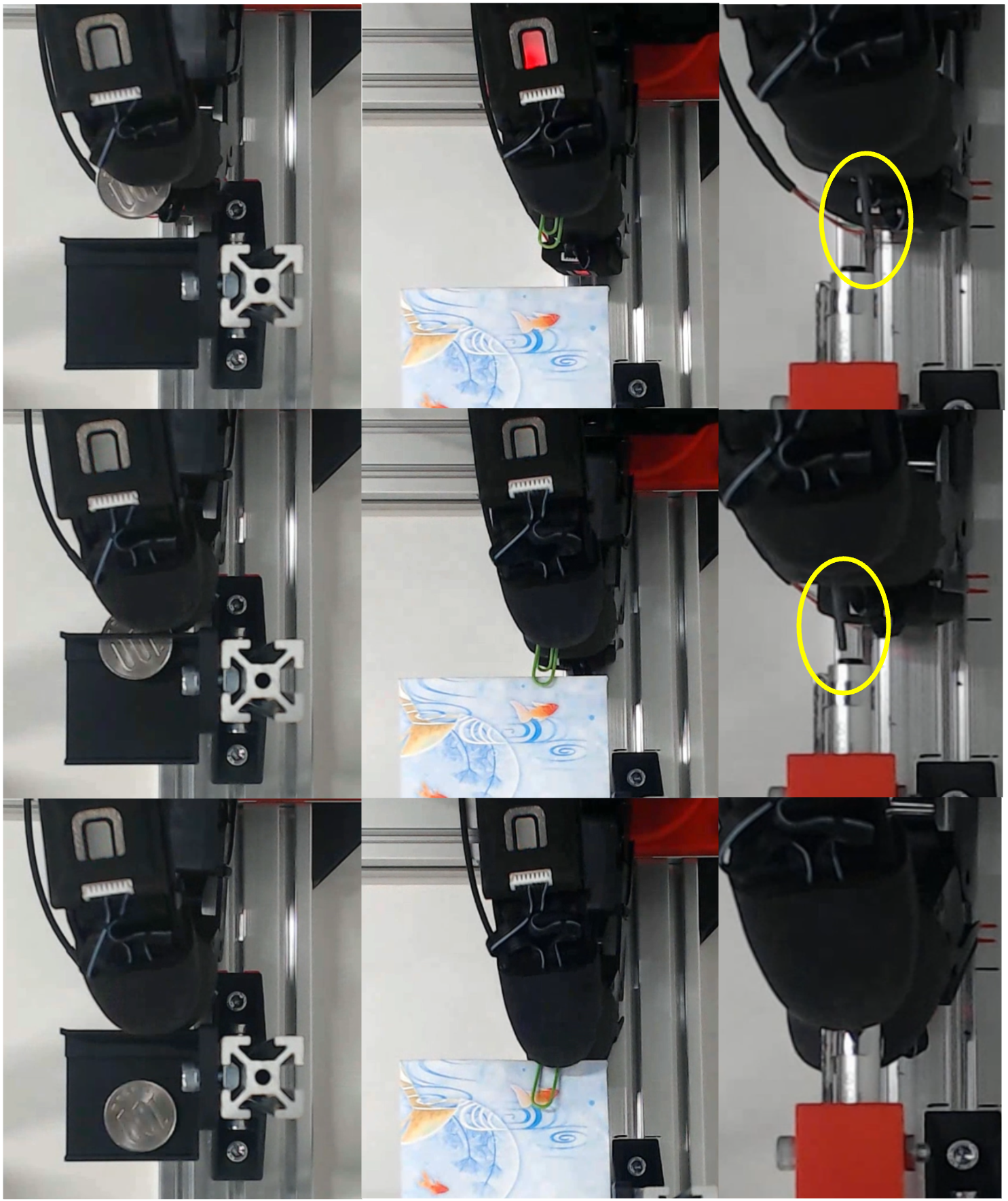}
    \caption{Diagrammatic representation of successful motion of the three tasks from initiation to completion.}
    \label{fig:tasks}
\end{figure}

\begin{table}[t]
\centering
\caption{Setting of Recurrent Neural Networks for Motion Learning}
\label{tab:rnn_settings}
\scriptsize
\renewcommand{\arraystretch}{1.2}
\resizebox{\columnwidth}{!}{%
\begin{tabular}{|p{2.0cm}|p{2.0cm}|p{2.0cm}|}
\hline
\textbf{Component} &
\textbf{RT + Self (st-rnn)} &
\textbf{RT Only (t-rnn)} \\
\hline
\textbf{Recurrent core} & \multicolumn{2}{c|}{RNN (LSTMCell), hidden dim $=100$} \\
\hline
\textbf{Inputs (size)} &
RT$+$ Self $+\,q_t$ \newline $180 + 180 + 8 = \mathbf{368}$ &
RT$+\,q_t$ \newline $180 + 8 = \mathbf{188}$ \\
\hline
\textbf{Decoder} &
\multicolumn{2}{c|}{Linear($100 \!\to\! 128$), ReLU; Linear($128 \!\to\! 196$)} \\
\hline
\textbf{Outputs (size)} &
\multicolumn{2}{c|}{(idx, thb): $90+90=180$, $q_{t+1}:8$, $q^{\mathrm{cmd}}_{t+1}:8$ $\Rightarrow$ total $\mathbf{196}$} \\
\hline
\end{tabular}
}
\end{table}

\section{Experiment}

The experimental evaluation is designed around four components: a self-touch prediction phase followed by three precision insertion tasks involving paper clip fixing, coin insertion, and pencil lead insertion, all, shown in Fig.~\ref{fig:tasks}. Each task was chosen to highlight the challenges of distinguishing self-touch (finger-finger contact) from external touch (object contact), where both produce small and ambiguous tactile signals. Success therefore requires accurate separation of these modalities.

\begin{figure}[ht]
    \centering
    \includegraphics[width=0.8\columnwidth]{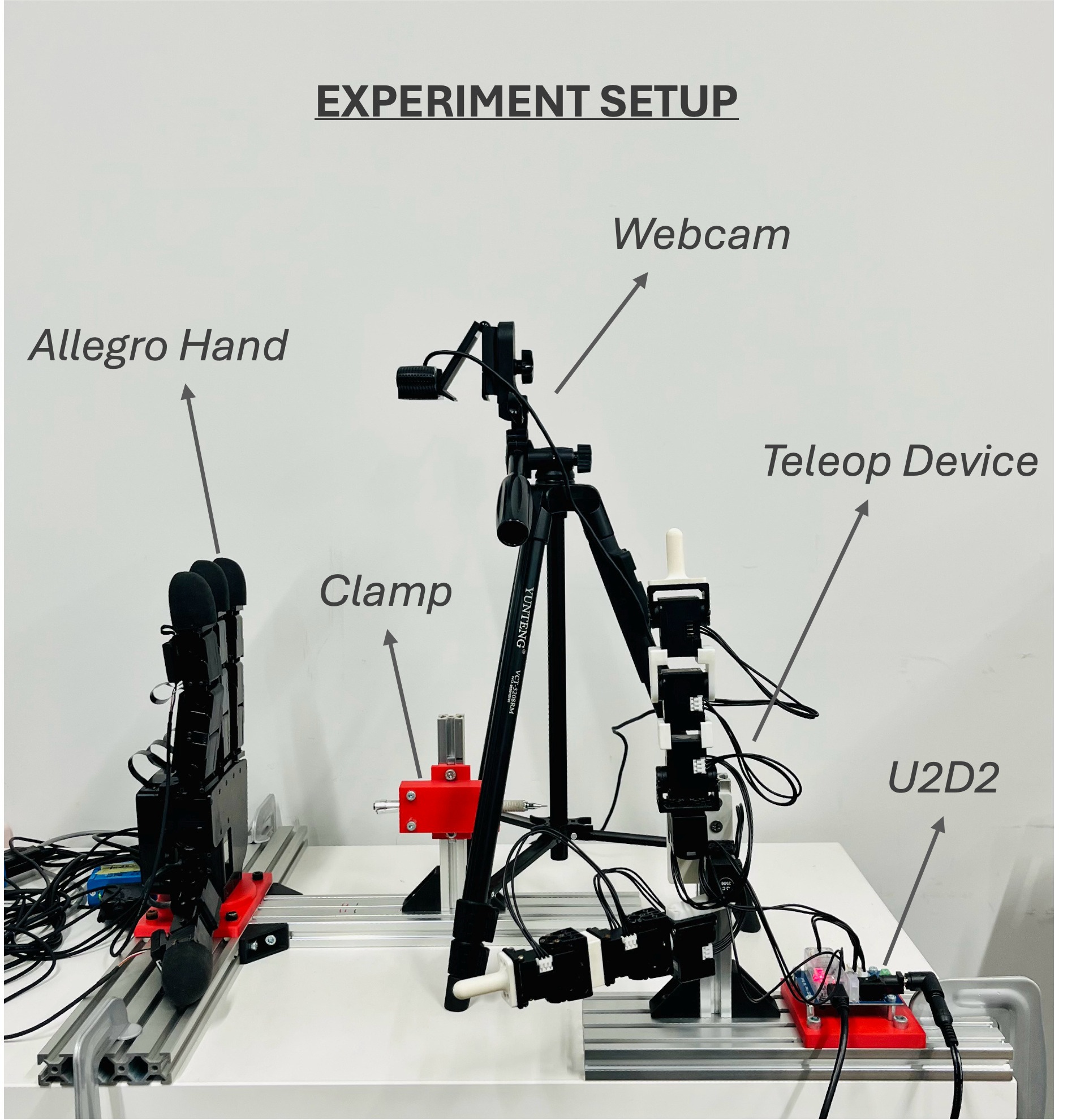}
    \caption{Diagrammatic representation of the setup: We teleoperate the Allegro Hand to collect data; the clamp holds the pencil for lead insertion and the webcam records motions.}
    \label{fig:setup}
\end{figure}

\begin{figure}[ht]
    \centering
    \includegraphics[width=1.0\columnwidth]{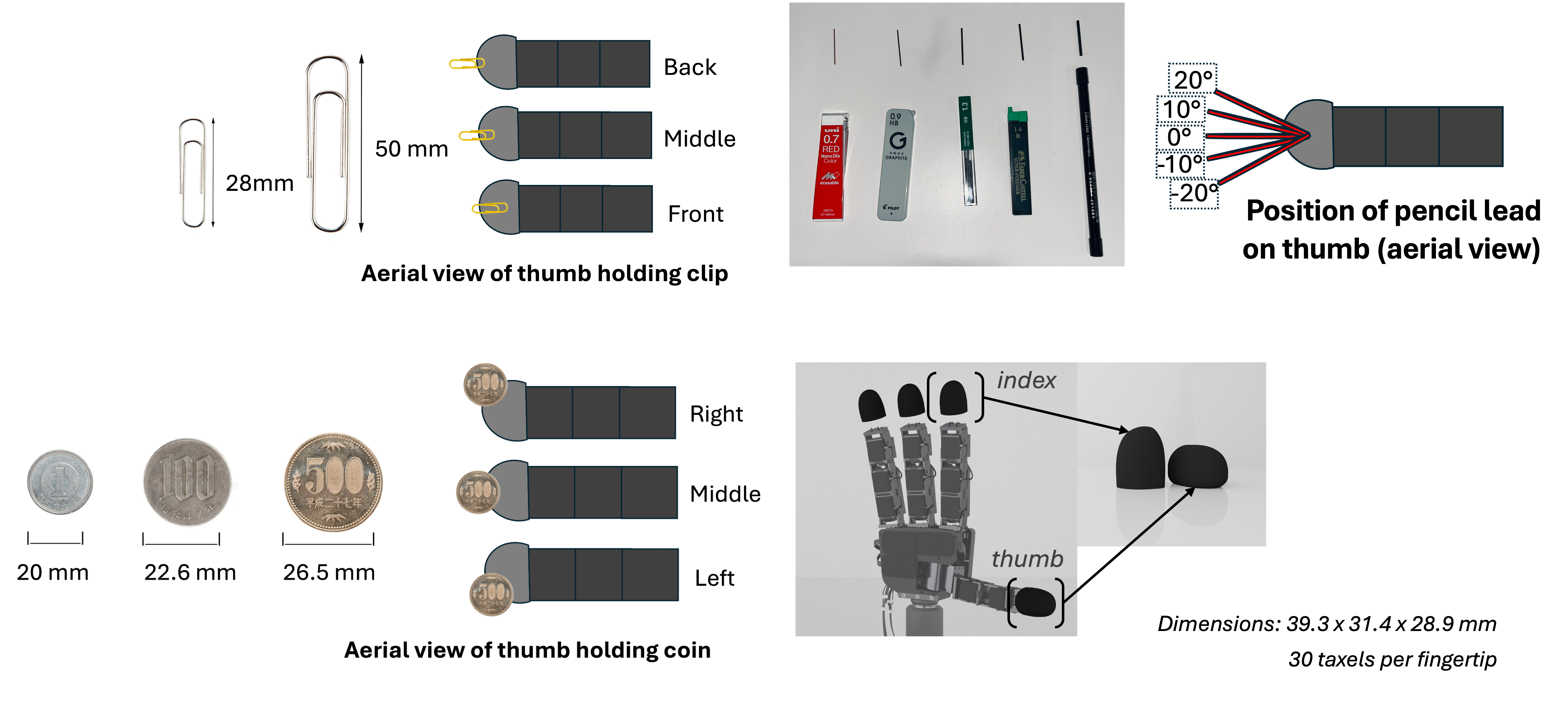}
    \caption{Diagrammatic representation of the material's size and orientations for the three insertion tasks.}
    \label{fig:positions}
\end{figure}

We employ the Allegro Hand, a four-fingered robotic hand from Wonik Robotics, with 16 degrees of freedom (DOF). The uSkin tactile sensors are provided by XELA Robotics: high-resolution tri-axial modules capable of measuring shear forces $(F_x, F_y)$ and normal force $(F_z)$ at each taxel.  The uSkin's individual taxels are spaced about 6.5 mm apart. Four uSCu curved fingertip sensors were mounted on the Allegro Hand, with experiments restricted to the index and thumb fingertips. A leader-follower teleoperation system was implemented to generate naturalistic motion trajectories, where a human operator manipulated a Dynamixel-based leader structure and the joint positions were transmitted in real time to the Allegro Hand via a U2D2 servo board from ROBOTIS e-shop, all, shown in Fig.~\ref{fig:setup}. The material's sizes along with the positions, orientations are shown in Fig.\ref{fig:positions}.

All three motion learning insertion tasks were trained with a batch size of 300, a learning rate of 0.001, and 6000 epochs. This combination of parameters provided stable gradient updates for the high-dimensional tactile inputs, thus allowing the optimizer to converge without divergence, and ensured sufficient training time to capture subtle patterns in contact-rich manipulation tasks.

\subsection{Self-Touch Prediction}
In this phase, the task is to predict tactile feedback that arises solely from self-touch, given the hand’s joint configuration. The objective is not to achieve a specific manipulation, but to ensure the model can anticipate the tactile patterns generated when the thumb and index finger interact without any object. Accurate prediction allows the model to understand self-generated signals.
To evaluate performance, we trained on 200 episodes, recorded at 10 Hz (400 time steps each) of random index–thumb motions that spanned a wide range of finger–finger interactions, from light grazing to sustained contact. This diversity of motion ensured that the learned model generalized beyond a single gesture or posture. Training was conducted with a batch size of 100, a learning rate of 0.001, and a total of 20,000 epochs. We used the Adam optimizer, and training was performed on an NVIDIA GeForce RTX 4070 GPU, which provided efficient convergence while maintaining stability in the learned predictions.

\begin{figure*}[ht]
  \centering
  \begin{minipage}[t]{0.49\textwidth}\centering
    \includegraphics[width=\linewidth]{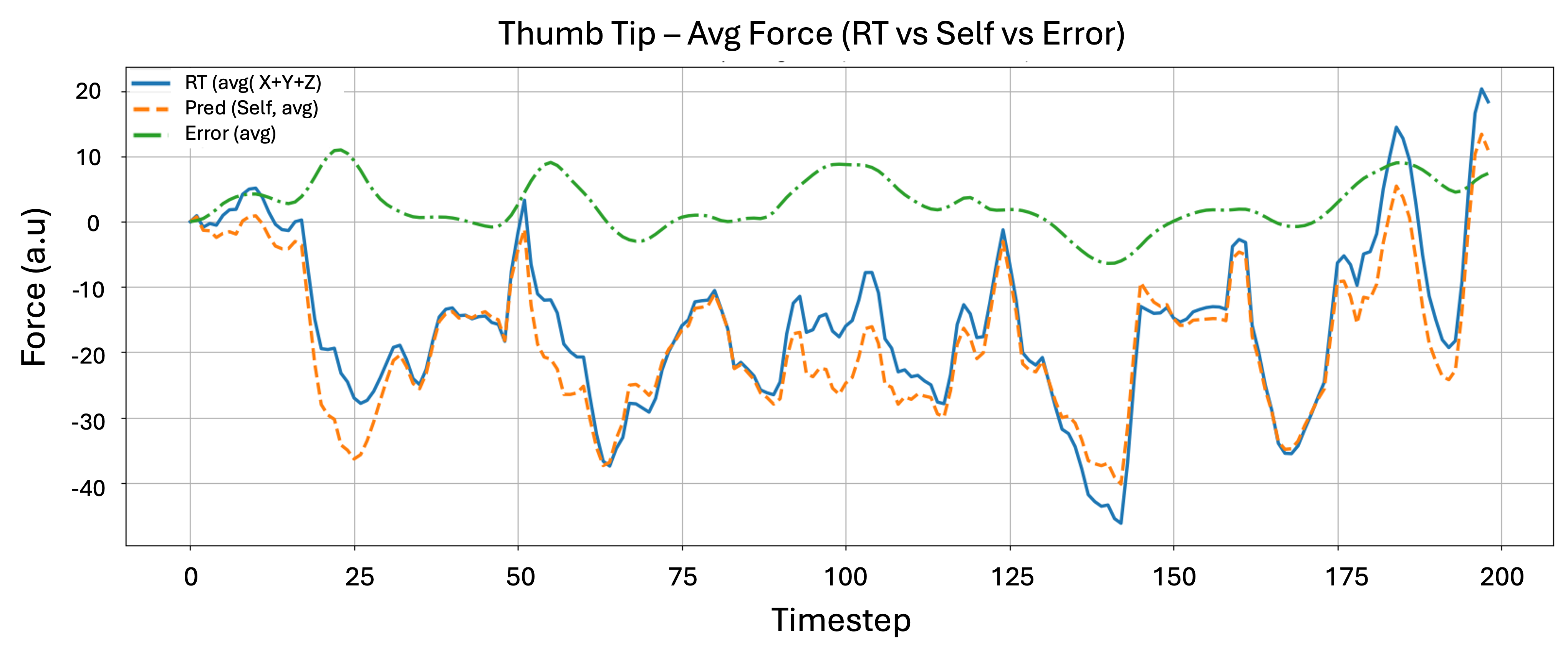}\\[-2pt]
    {\scriptsize (a) \textbf{Thumb Tip} — Avg force: RT, Self, Error}
  \end{minipage}\hfill
  \begin{minipage}[t]{0.49\textwidth}\centering
    \includegraphics[width=\linewidth]{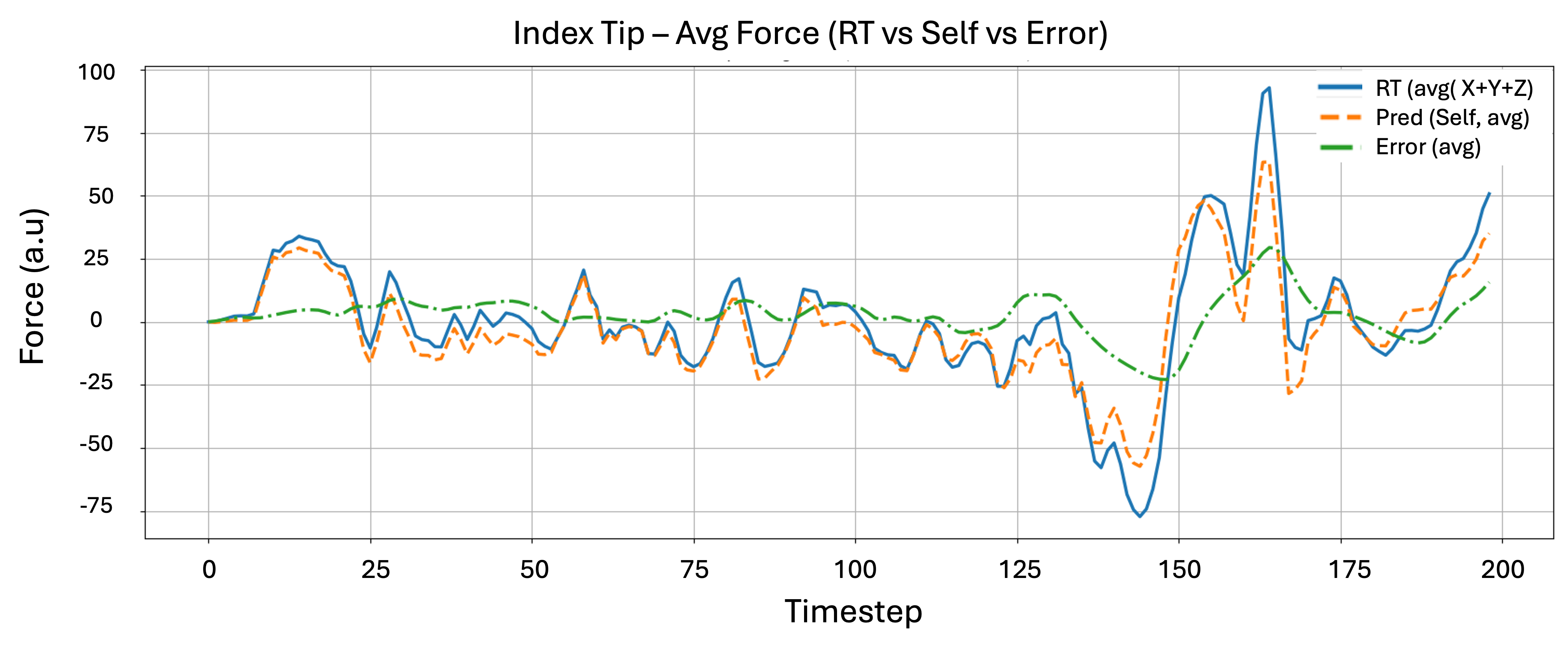}\\[-2pt]
    {\scriptsize (b) \textbf{Index Tip} — Avg force: RT, Self, Error}
  \end{minipage}
  \caption{Self-touch prediction on \emph{test} episodes. During sustained finger–finger contact the error stays near zero; spikes appear around rapid transitions and occasional random interference, which could be desirable for control because they highlight unexpected changes or events rather than continuous predictable contact.}
  \label{fig:selftouch_avg}
\end{figure*}

\subsection{Paper Clip Fixing}
In the paper clip fixing task, two clip sizes (small and big) were placed at three lateral positions: \{front, middle, back\}. Training was performed on the \{front, back\} positions, while the \{middle\} position was reserved for testing. Two types of paper clips were employed: a large clip measuring 50 mm in length and a small clip of 28 mm.

\begin{table}[ht]
\centering
\caption{Distribution of experimental task for paper clip fixing}
\begin{tabular}{c|c|c|c}
\hline
Clip Size & Front & Middle & Back \\
\hline
Small & Train & Test & Train \\
Big   & Train & Test & Train \\
\hline
\end{tabular}
\label{tab:clip}
\end{table}

\subsection{Coin Insertion}
The coin insertion task tested three denominations (1 yen, 100 yen, 500 yen), each inserted into a narrow slot at three lateral positions: \{left, middle, right\}. Training used \{left, right\}, while \{middle\} was reserved for testing. For coin-based tasks, three denominations common in Japan were considered: the 1-yen coin (20 mm diameter, 1.5 mm thickness), the 100-yen coin (22.6 mm diameter, 1.7 mm thickness), and the 500-yen coin (26.5 mm diameter, 1.8 mm thickness). For the task itself, a 3D-printed coin slot was modeled on Japanese vending machine specifications, with a slot length of 32 mm and a width of 2 mm.

\begin{table}[ht]
\centering
\caption{Distribution of experimental task for coin insertion}
\begin{tabular}{c|c|c|c}
\hline
Coin Type & Left & Middle & Right \\
\hline
1 yen   & Train & Test & Train \\
100 yen & Train & Test & Train \\
500 yen & Train & Test & Train \\
\hline
\end{tabular}
\label{tab:coin}
\end{table}

\subsection{Pencil Lead Insertion}
Pencil leads of five diameters 
\(\{0.7, 0.9, 1.3, 1.4, 2.0\ \text{mm}\}\) 
were inserted at five angles relative to the thumb’s surface 
\(\{-20^\circ, -10^\circ, 0^\circ, +10^\circ, +20^\circ\}\). Training was performed on the extreme angles \{-20°, 0°, +20°\}, with testing on intermediate angles \{-10°, +10°\}, probing angular interpolation. For pencil-lead insertion tasks, HB leads of 0.7, 0.9, 1.3, 1.4, and 2.0 mm diameters were inserted into the back end of a mechanical drafting pencil, which had an outer opening diameter of 3.0 mm.

\begin{table}[ht]
\centering
\caption{Distribution of experimental task for pencil lead insertion}
\begin{tabular}{c|c|c|c|c|c}
\hline
Diameter (mm) & -20$^\circ$ & -10$^\circ$ & 0$^\circ$ & +10$^\circ$ & +20$^\circ$ \\
\hline
0.7 & Train & Test & Train & Test & Train \\
0.9 & Test  & Test & Test  & Test & Test  \\
1.3 & Train & Test & Train & Test & Train \\
1.4 & Test  & Test & Test  & Test & Test  \\
2.0 & Train & Test & Train & Test & Train \\
\hline
\end{tabular}
\label{tab:pencil}
\end{table}

\section{Evaluation}

We evaluate at two levels: (i) the accuracy of self-touch prediction (isolation of self contact) and (ii) task-level performance on three precision manipulations with train–test splits designed to probe generalization. Metrics are per-condition success rates (10 trials each) and qualitative feature-space analyses (PCA; reported separately). Totals in each table are reported as successes / trials and the corresponding percentage.

\subsection{Self-Touch Prediction}

Across held-out test episodes, the predicted self-touch signal aligns closely with the raw-tactile (RT) trace for both fingers. For the thumb tip (Fig.~\ref{fig:selftouch_avg}a), the prediction overlaps the RT curve over long continuous segments, including during rapid excursions, with the error channel staying within a few arbitrary units (a.u.). Small bumps appear around force transitions, typically $8$–$10$,a.u., which are much smaller than the typical self-contact magnitude of $40$–$50$,a.u. Rather than being harmful, these localized deviations can be beneficial because they flag moments of rapid change that the controller may need to monitor. For the index tip (Fig.~\ref{fig:selftouch_avg}b), the prediction follows the RT trace tightly at moderate amplitudes, with the error channel remaining essentially flat except for two brief intervals where RT rises sharply. Even the largest deviation, peaking at $25$–$30$,a.u.\ during a transient re-grasp event, is well below the peak self-contact magnitude of $60$–$70$,a.u. Overall, the error behaves as intended: it remains near zero during finger–finger contact, while non-zero responses appear primarily at contact onsets, offsets. High correlations on self-contact segments ($r \approx 0.96$ for thumb, $r \approx 0.98$ for index) confirm that the model effectively cancels self-touch.

\subsection{Paper Clip Fixing}
Table~\ref{tab:clip_results_icra} shows success rates for two clip sizes across three placements. With raw tactile (RT) only, overall success is \(70\%\) (42/60). Combining RT with Self improves to \(95\%\) (57/60). The \emph{front} placement is the most difficult because the fingers sometimes push the clip only partway onto the paper; in these cases, the raw tactile signal can resemble a fully inserted state and be misclassified as successful completion. Incorporating RT+Self reduces such errors by disambiguating finger–finger contact from actual clip–paper contact. The unseen \emph{middle} placement (test) attains \(100\%\) success under RT+Self for both sizes.

\begin{table}[h!]
  \centering
  \footnotesize
  \setlength{\tabcolsep}{5pt}
  \caption{Paper clip fixing success rates (10 trials per condition).}
  \label{tab:clip_results_icra}
  \begin{tabular}{lcc|cc}
    \hline
    & \multicolumn{2}{c|}{RT only} & \multicolumn{2}{c}{RT+Self} \\
    \cline{2-5}
    & Big & Small & Big & Small \\
    \hline
    Back          & 8/10 & 7/10 & 10/10 & 10/10 \\
    Middle (test) & 8/10 & 8/10 & 10/10 & 10/10 \\
    Front         & 6/10 & 5/10 & 9/10  & 8/10  \\
    \hline
    Total         & \multicolumn{2}{c|}{42/60 = 70\%} & \multicolumn{2}{c}{57/60 = 95\%} \\
    \hline
  \end{tabular}
\end{table}

\subsection{Coin Insertion}
Table~\ref{tab:coin_results_icra} summarizes results for three coin sizes and three lateral slots. RT only achieves \(68\%\) (61/90), while RT+Self reaches \(92\%\) (83/90). Failures concentrate at the \emph{left} slot for larger coins (100\,yen, 500\,yen), where edges are frequently grazed. RT+Self improves alignment and generalization: the unseen \emph{middle} slot (test) is \(100\%\) for all coin sizes.

\begin{table}[h!]
  \centering
  \footnotesize
  \setlength{\tabcolsep}{5pt}
  \caption{Coin insertion success rates (10 trials per condition).}
  \label{tab:coin_results_icra}
  \begin{tabular}{lccc|ccc}
    \hline
    & \multicolumn{3}{c|}{RT only} & \multicolumn{3}{c}{RT+Self} \\
    \cline{2-7}
    & 1\,yen & 100\,yen & 500\,yen & 1\,yen & 100\,yen & 500\,yen \\
    \hline
    Left          & 6/10 & 7/10 & 7/10 & 9/10  & 9/10  & 7/10 \\
    Middle (test) & 7/10 & 7/10 & 7/10 & 10/10 & 10/10 & 10/10 \\
    Right         & 7/10 & 7/10 & 6/10 & 10/10 & 9/10  & 9/10 \\
    \hline
    Total         & \multicolumn{3}{c|}{61/90 = 68\%} & \multicolumn{3}{c}{83/90 = 92\%} \\
    \hline
  \end{tabular}
\end{table}

\subsection{Pencil-Lead Insertion}
Lead insertion is the most demanding setting (Table~\ref{tab:lead_results_icra}). Overall success is \(26\%\) (66/250) for RT only, and improves to \(58\%\) (146/250) with RT+Self. Larger diameters (1.3\,mm, 2.0\,mm) succeed more often than thin leads (0.7\,mm, 0.9\,mm) due to stronger tactile signatures and reduced sensitivity to slight misalignment. The success rate lowers at both test (\(-10^\circ\), \(+10^\circ\)) and trained angles (\(-20^\circ\), \(+20^\circ\)), since precision is difficult and some trials hit or slid along the pencil edge. Nonetheless, RT+Self improves both trained and test angles, showing better robustness to fine geometric constraints.

\begin{table}[h!]
  \centering
  \footnotesize
  \setlength{\tabcolsep}{4pt}
  \caption{Pencil-lead insertion success rates (10 trials per condition). (t) denotes test splits.}
  \label{tab:lead_results_icra}
  \begin{tabular}{lccccc}
    \hline
    & \multicolumn{5}{c}{RT only} \\
    \cline{2-6}
    & 0.7 & 0.9\,(t) & 1.3 & 1.4\,(t) & 2.0 \\
    \hline
    \(-20^\circ\)    & 0/10 & 1/10 & 3/10 & 3/10 & 3/10 \\
    \(-10^\circ\)(t) & 1/10 & 2/10 & 3/10 & 3/10 & 4/10 \\
    \(0^\circ\)      & 2/10 & 2/10 & 4/10 & 4/10 & 5/10 \\
    \(+10^\circ\)(t) & 1/10 & 3/10 & 3/10 & 3/10 & 5/10 \\
    \(+20^\circ\)    & 1/10 & 2/10 & 2/10 & 2/10 & 4/10 \\
    \hline
    Total & \multicolumn{5}{c}{66/250 = 26\%} \\
    \hline
  \end{tabular}

  \begin{tabular}{lccccc}
    \hline
    & \multicolumn{5}{c}{RT+Self} \\
    \cline{2-6}
    & 0.7 & 0.9\,(t) & 1.3 & 1.4\,(t) & 2.0 \\
    \hline
    \(-20^\circ\)    & 3/10 & 4/10 & 6/10 & 6/10 & 7/10 \\
    \(-10^\circ\)(t) & 5/10 & 6/10 & 7/10 & 7/10 & 8/10 \\
    \(0^\circ\)      & 5/10 & 5/10 & 8/10 & 8/10 & 9/10 \\
    \(+10^\circ\)(t) & 4/10 & 4/10 & 6/10 & 7/10 & 8/10 \\
    \(+20^\circ\)    & 4/10 & 3/10 & 5/10 & 5/10 & 6/10 \\
    \hline
    Total & \multicolumn{5}{c}{146/250 = 58\%} \\
    \hline
  \end{tabular}
\end{table}

\begin{figure*}[ht]
  \centering
  % -------- Column 1: Paper Clip --------
  \begin{minipage}[t]{0.31\textwidth}
    \centering
    \includegraphics[width=\linewidth]{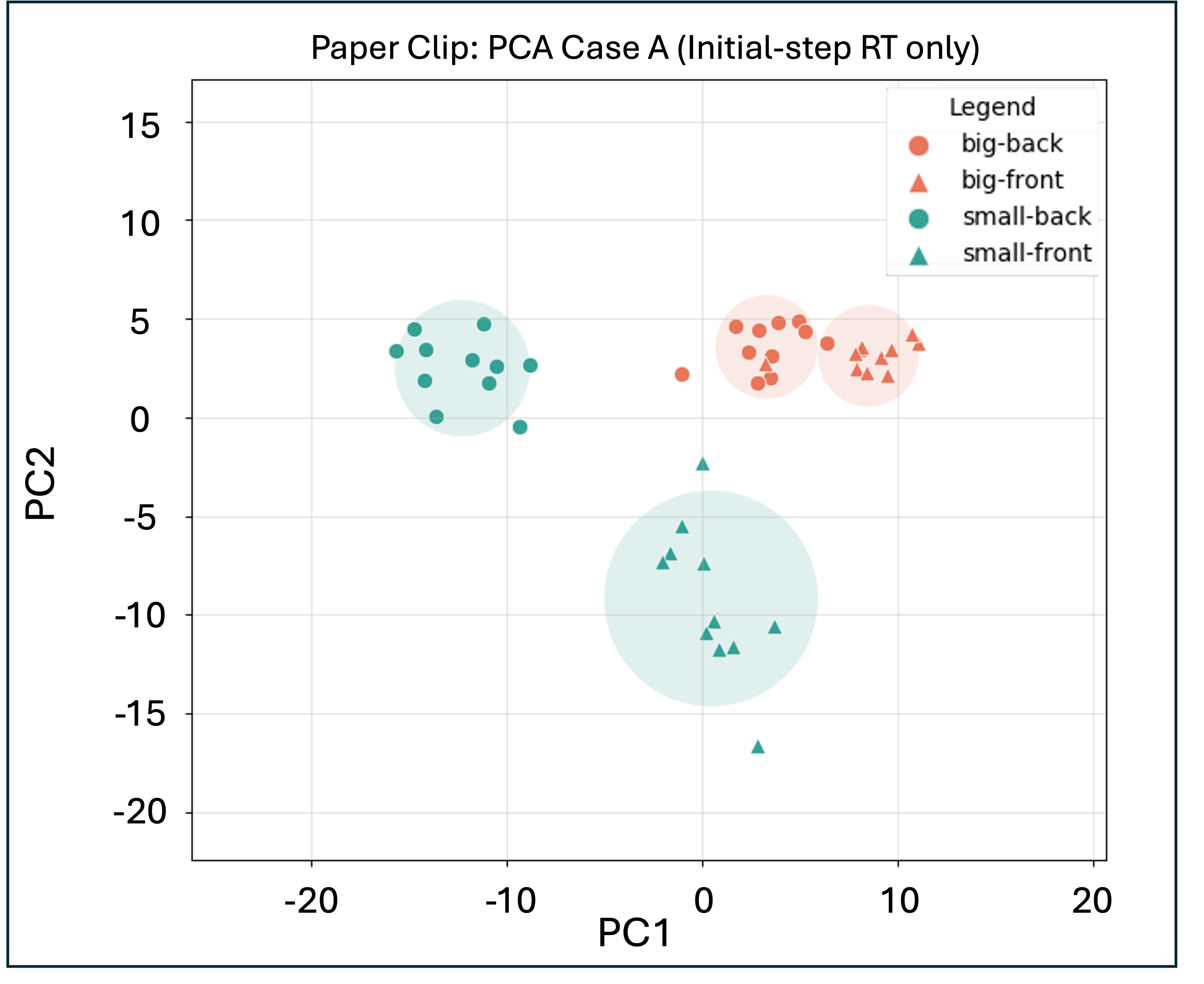}\\[-2pt]
    {\footnotesize (a) Paper Clip — Case A (RT)}
  \end{minipage}\hfill
  \begin{minipage}[t]{0.31\textwidth}
    \centering
    \includegraphics[width=\linewidth]{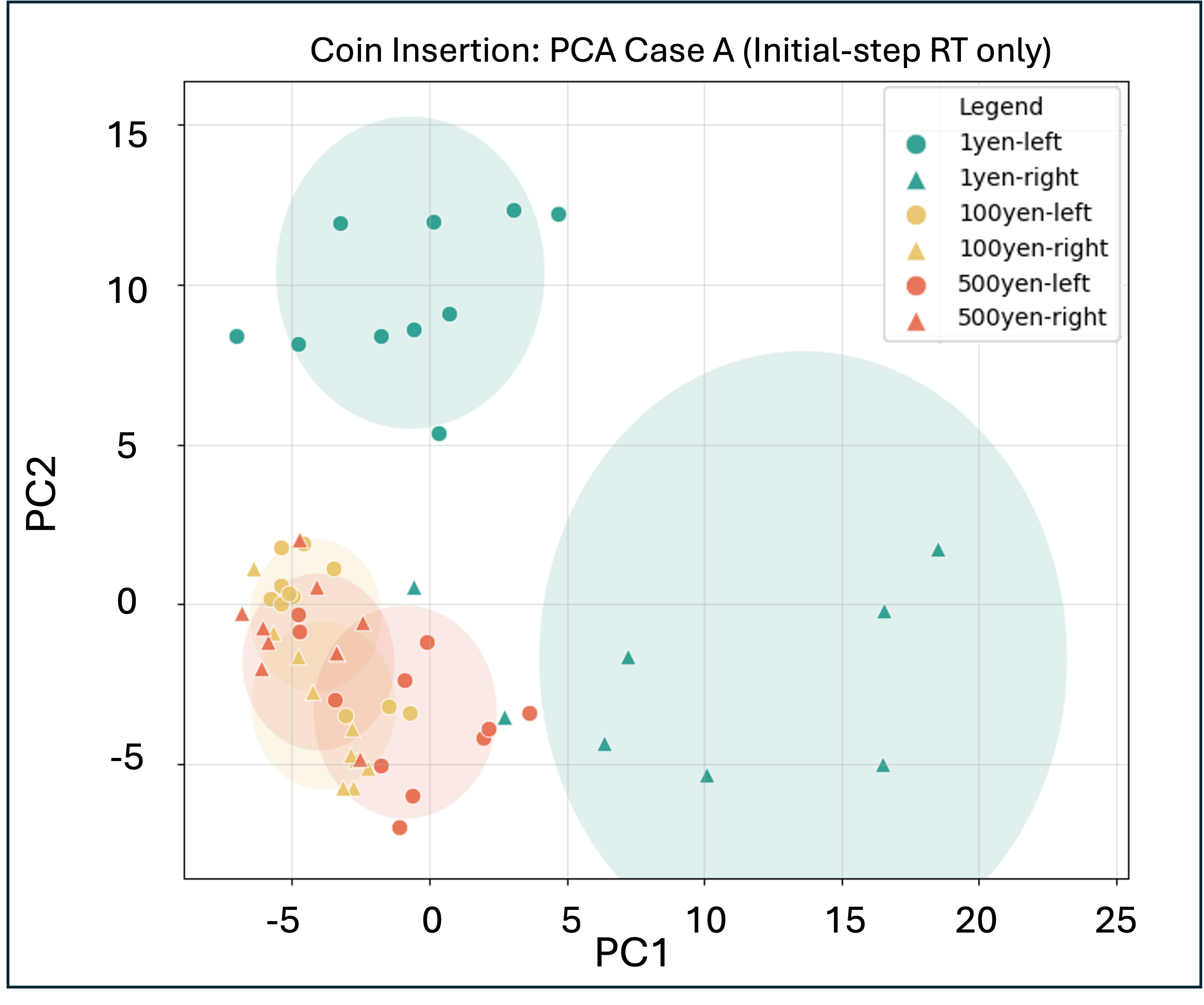}\\[-2pt]
    {\footnotesize (c) Coins — Case A (RT)}
  \end{minipage}\hfill
  \begin{minipage}[t]{0.31\textwidth}
    \centering
    \includegraphics[width=\linewidth]{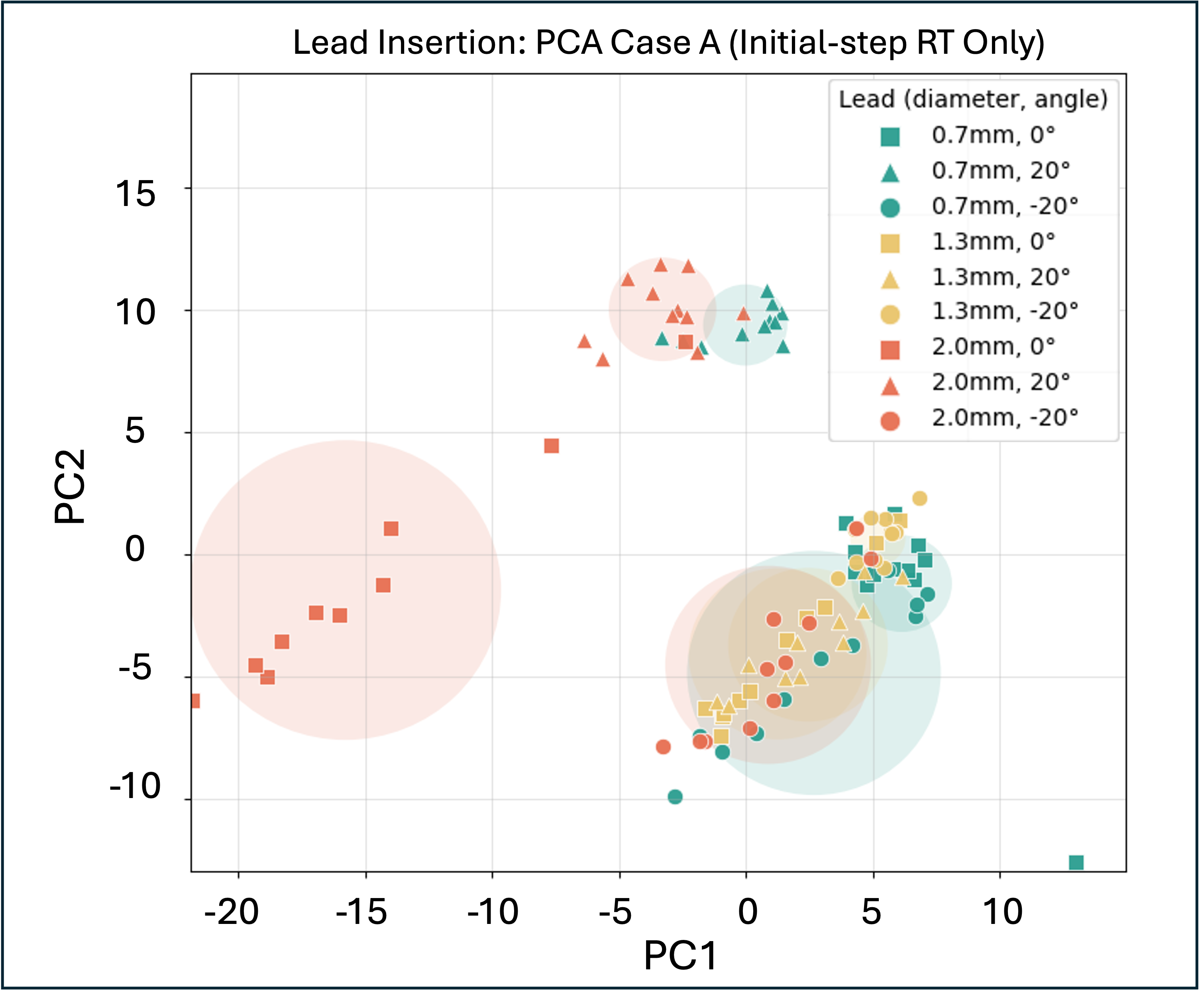}\\[-2pt]
    {\footnotesize (e) Lead — Case A (RT)}
  \end{minipage}

  % -------- Column 2: Case B panels --------
  \begin{minipage}[t]{0.31\textwidth}
    \centering
    \includegraphics[width=\linewidth]{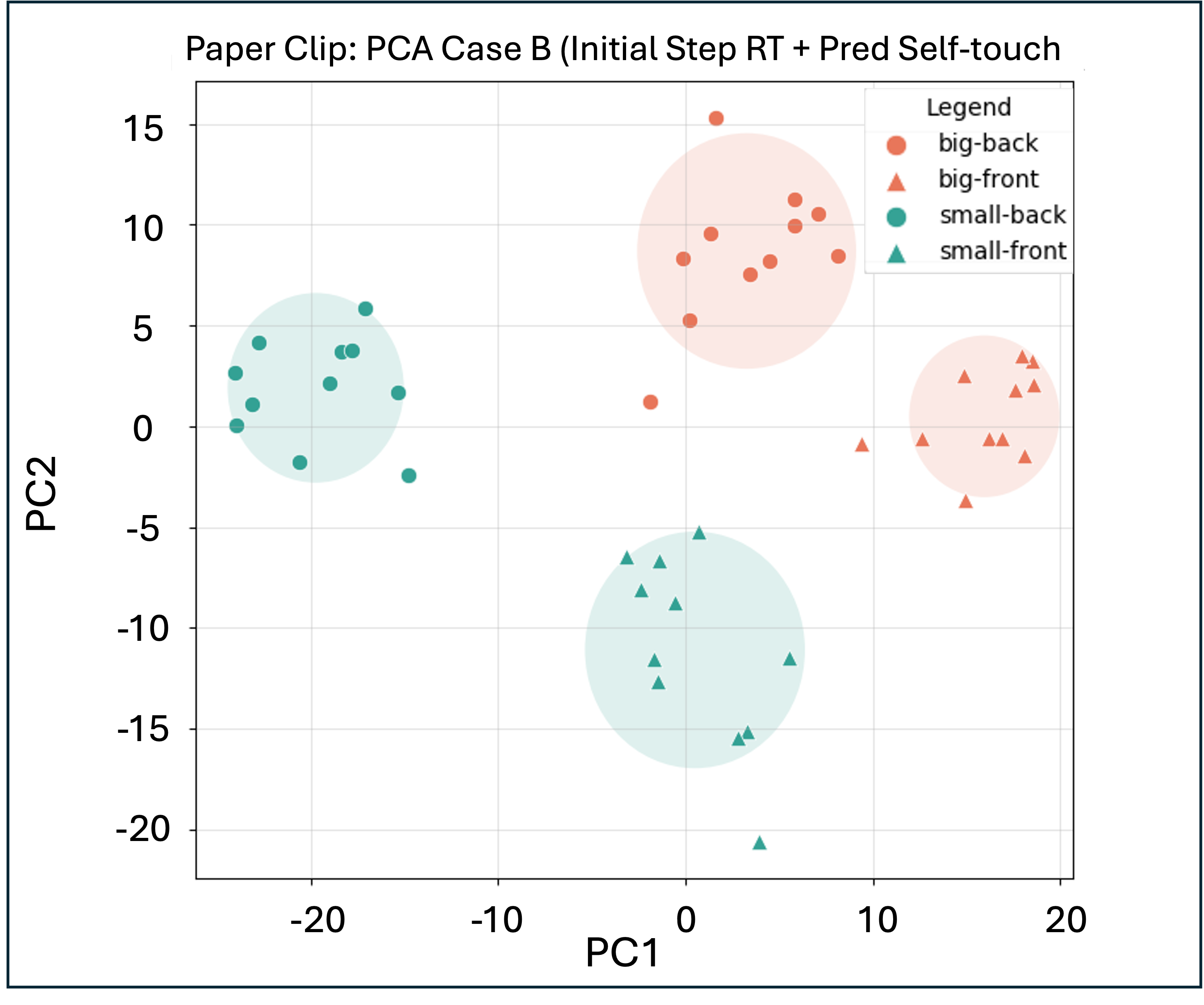}\\[-2pt]
    {\footnotesize (b) Paper Clip — Case B (RT$\oplus$ Self)}
  \end{minipage}\hfill
  \begin{minipage}[t]{0.31\textwidth}
    \centering
    \includegraphics[width=\linewidth]{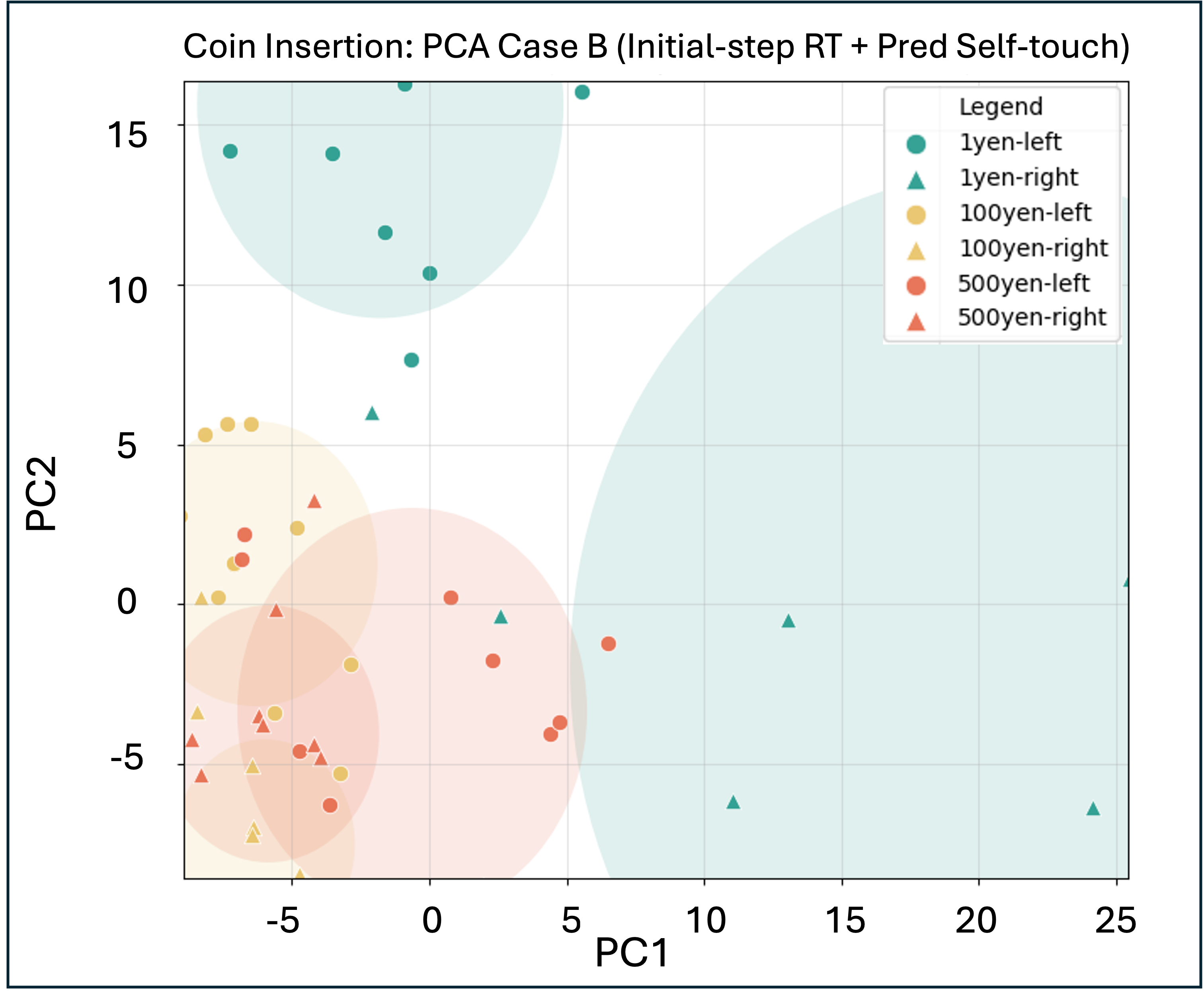}\\[-2pt]
    {\footnotesize (d) Coins — Case B (RT$\oplus$ Self)}
  \end{minipage}\hfill
  \begin{minipage}[t]{0.31\textwidth}
    \centering
    \includegraphics[width=\linewidth]{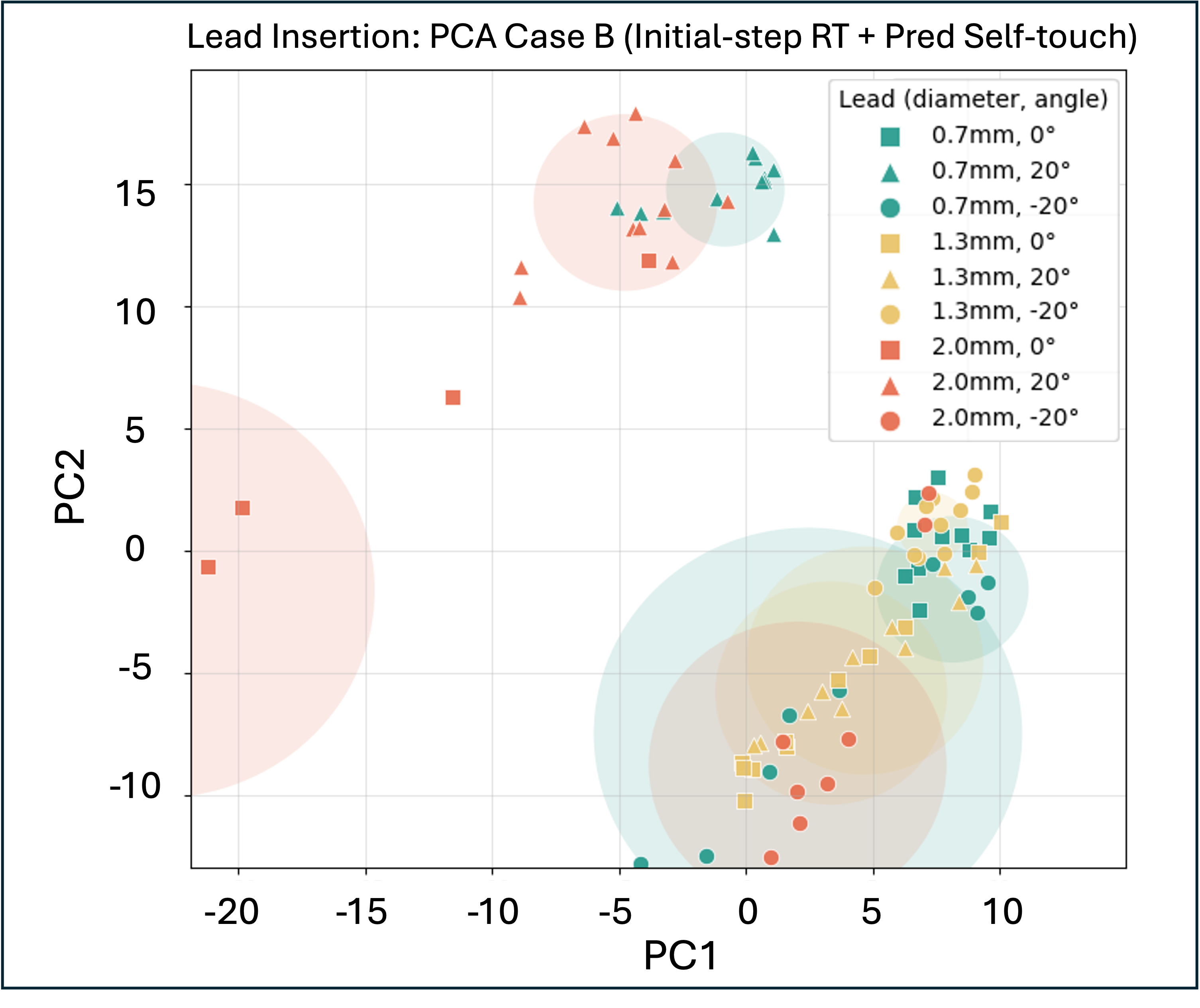}\\[-2pt]
    {\footnotesize (f) Lead — Case B (RT$\oplus$ Self)}
  \end{minipage}

  \caption{PCA feature–space comparison at the initial step for three tasks (paper clip, coins, lead) under two conditions: (Case A) RAW RT and (Case B) RT augmented with predicted self-touch. Each column corresponds to one task, with the top row showing Case A and the bottom row showing Case B. Each point is a trial; translucent disks indicate within-class spread.}
  \label{fig:pca_all}
\end{figure*}

\subsection{PCA Feature–Space Analysis (Case A vs.\ Case B)}

Paper Clip Fixing \emph{(Fig.~8(a) and 8(b))}: In Case~A (RT), the initial tactile features are strongly influenced by grasp preload and finger–finger shear. As a result, trials from the back and front placements form clusters that are somewhat distinct but still overlap, indicating that self-touch effects obscure placement-specific differences. In Case~B, incorporating the self-touch prediction reduces the influence of self-generated signals, which lowers variance along the axis aligned with finger–finger contacts and increases the inter-centroid distance between front and back for each size. Mechanically, the front placement induces finger–finger contact that mimics clip–paper completion; incorporating self-touch mitigates this, yielding cleaner boundaries.

Coin Insertion \emph{(Fig.~8(c) and 8(d))}: Although the pre-insertion state is similar across the train and test slots, adding the self-touch prediction yields a \emph{stabilized} embedding: clusters translate coherently and several classes tighten (notably 1\,yen and 500\,yen right), indicating reduced nuisance variance from finger–finger preload. The \emph{100\,yen} and \emph{500\,yen} groups still partially merge—their thicknesses differ by only $\approx 0.1\,\mathrm{mm}$, so RT only produces near-identical normal forces and contact-patch areas—but in Case~B the covariance shrinks and spacing increases, making slot boundaries less entangled. This representation already supports better alignment and generalization: RT+Self attains improved results on the unseen middle slot across all coin sizes.

Lead Insertion \emph{(Fig.~8(e) and 8(f))}: In CaseA, the diameter pattern is already visible, but the \emph{0.7 mm} and \emph{1.3 mm} groups partly overlap. Their small contact areas and weak normal forces give low signal quality, and even slight angle changes ($\pm20^\circ$) alter the shear direction enough to blur diameter differences. Case B keeps the diameter separation clearer and also shows some angle-based grouping, but the thinnest leads are still the hardest to tell apart. This explains why success is lower at larger angles and with very thin leads.

\section{Conclusion}
We presented \textbf{TaSA}, a two-phased deep predictive framework that learns a self-touch model, where the error channel signal remained almost near zero during pure finger–finger contact and thus enabled a stable understanding of the external object contact for control. Across all three precision tasks; TaSA consistently improved robustness in contact-rich settings: it improved the detection capabilities for clips, generalized to an unseen coin slot with perfect success, and provided the largest gains for thin leads and larger angles. Analysis of PCA embeddings confirmed the intended behavior: class geometry became clearer once self-touch was attenuated.

As a next step, this framework could be extended from two-finger settings to multi-fingered manipulation. Multi-finger manipulation tasks introduce more frequent finger–finger and finger–palm interactions, which are often disregarded in existing approaches. By predicting these structured self-touch patterns, TaSA could help robots disambiguate self- and object-generated forces, a distinction that is critical for stable grasps, coordinated re-grasps, and fine in-hand adjustments. This extension represents the natural higher-DOF case, where preserving object integrity requires accurate handling of dense self-contacts. Moreover, we tried to extract external touch as a separate input from the raw-tactile and preliminary training for the lead insertion yields promise, yet further investigation is required and will be propagated in the future. 

%%%%%%%%%%%%%%%%%%%%%%%%%%%%%%%%%%%%%%%%%%%%%%%%%%%%%%%%%%%%%%%%%%%%%%%%%%%%%%%%
\addtolength{\textheight}{-12cm}   % This command serves to balance the column lengths

%%%%%%%%%%%%%%%%%%%%%%%%%%%%%%%%%%%%%%%%%%%%%%%%%%%%%%%%%%%%%%%%%%%%%%%%%%%%%%%%


\begin{thebibliography}{99}

\bibitem{1}
K. Kilteni and H. H. Ehrsson, 
``Predictive attenuation of touch and tactile gating are distinct perceptual phenomena,'' 
\emph{iScience}, vol. 25, no. 4, p. 104077, 2022, doi: 10.1016/j.isci.2022.104077.

\bibitem{2}
K. Kilteni, C. Houborg, and H. H. Ehrsson, 
``Brief temporal perturbations in somatosensory reafference disrupt perceptual and neural attenuation and increase supplementary motor area--cerebellar connectivity,'' 
\emph{Journal of Neuroscience}, vol. 43, no. 28, pp. 5251--5263, 2023, doi: 10.1523/JNEUROSCI.2289-22.2023.

\bibitem{3}
K. Kilteni and H. H. Ehrsson, 
``Body ownership determines the attenuation of self-generated tactile sensations,'' 
\emph{Proceedings of the National Academy of Sciences}, vol. 114, no. 31, pp. 8426--8431, 2017, doi: 10.1073/pnas.1703347114.

\bibitem{4}
S. Funabashi, S. Ogasa, T. Isobe, T. Ogata, A. Schmitz, T. P. Tomo, and S. Sugano, 
``Variable In-Hand Manipulations for Tactile-Driven Robot Hand via CNN-LSTM,'' 
in \emph{Proc. IEEE/RSJ Int. Conf. Intelligent Robots and Systems (IROS)}, 2019, pp. 7445--7452, doi: 10.1109/IROS40897.2019.8968073.

\bibitem{5}
S. Funabashi, T. Isobe, H. Fei, A. Hiramoto, A. Schmitz, S. Sugano, and T. Ogata, 
``Multi-Fingered In-Hand Manipulation with Various Object Properties Using Graph Convolutional Networks and Distributed Tactile Sensors,'' 
in \emph{Proc. IEEE/RSJ Int. Conf. Intelligent Robots and Systems (IROS)}, 2020, pp. 9380--9386, doi: 10.1109/IROS45743.2020.9341389.

\bibitem{6}
I.~Guzey, Y.~Dai, B.~Evans, S.~Chintala, and L.~Pinto, 
``See to Touch: Learning Tactile Dexterity through Visual Incentives,'' 
in \emph{Proc. IEEE Int. Conf. Robotics and Automation (ICRA)}, 
Yokohama, Japan, 2024, pp.~13825--13832, 
doi: 10.1109/ICRA57147.2024.10611407.

\bibitem{7}
I.~Guzey, B.~Evans, S.~Chintala, and L.~Pinto, 
``Dexterity from Touch: Self-Supervised Pre-Training of Tactile Representations with Robotic Play,'' 
\emph{arXiv preprint} arXiv:2303.12076, 2023.

\bibitem{8}
T. Ueno, Y. Takeda, T. Iwasaki, K. Kutsuzawa, A. Schmitz, S. Sugano, and T. Ogata, 
``Multi-Fingered Dragging of Unknown Objects and Orientations Using Distributed Tactile Information Through Vision-Transformer and LSTM,'' 
in \emph{Proc. IEEE/RSJ Int. Conf. Intelligent Robots and Systems (IROS)}, Abu Dhabi, 2024, pp. 7445--7452, doi: 10.1109/IROS58592.2024.10802283.

\bibitem{9}
M. Hara, P. Pozeg, G. Rognini, K. Fukuhara, A. Yamamoto, T. Higuchi, O. Blanke, and R. Salomon, 
``Voluntary self-touch increases body ownership,'' 
\emph{Frontiers in Psychology}, vol. 6, p. 1509, 2015, doi: 10.3389/fpsyg.2015.01509.

\bibitem{10} 
P. Lanillos and G. Cheng, 
``Adaptive robot body learning and estimation through predictive coding,'' 
in \emph{Proc. IEEE/RSJ Int. Conf. Intelligent Robots and Systems (IROS)}, Madrid, Spain, 2018, pp. 4083--4090.

\bibitem{11} 
P. D. H. Nguyen, Y. K. Georgie, E. Kayhan, M. Eppe, V. V. Hafner, and S. Wermter, 
``Sensorimotor representation learning for an 'active self' in robots: A model survey,'' 
\emph{arXiv preprint arXiv:2011.12860}, 2021.

\bibitem{12} 
F. Gama, M. Shcherban, M. Rolf, and M. Hoffmann, 
``Goal-directed tactile exploration for body model learning through self-touch on a humanoid robot,'' 
\emph{IEEE Transactions on Cognitive and Developmental Systems}, vol. 15, no. 2, pp. 419--433, 2023.

\bibitem{13}
O.~Azulay, D.~M.~Ramesh, N.~Curtis, and A.~Sintov, 
``Visuotactile-Based Learning for Insertion With Compliant Hands,'' 
\emph{IEEE Robotics and Automation Letters}, vol.~10, no.~4, pp.~4053--4060, Apr.~2025, 
doi: 10.1109/LRA.2025.3549657.

\bibitem{14}
K.~Nozu and K.~Shimonomura, 
``Robotic Bolt Insertion and Tightening Based on In-Hand Object Localization and Force Sensing,'' 
in \emph{Proc. IEEE/ASME Int. Conf. on Advanced Intelligent Mechatronics (AIM)}, 
Auckland, New Zealand, 2018, pp.~310--315, 
doi: 10.1109/AIM.2018.8452338.

\bibitem{15}
K.~Miyama, S.~Hasegawa, K.~Kawaharazuka, N.~Yamaguchi, K.~Okada, and M.~Inaba, 
``Design of a Five-Fingered Hand with Full-Fingered Tactile Sensors Using Conductive Filaments and Its Application to Bending after Insertion Motion,'' 
in \emph{Proc. IEEE-RAS Int. Conf. on Humanoid Robots (Humanoids)}, 
Ginowan, Japan, 2022, pp.~780--785, 
doi: 10.1109/Humanoids53995.2022.10000181.

\bibitem{16}
Z.~Gong, Z.~Gao, G.~Zhu, and T.~Zhang, 
``A Solid-Liquid Composite Flexible Bionic Three-Axis Tactile Sensor for Dexterous Hands,'' 
\emph{IEEE Robotics and Automation Letters}, vol.~9, no.~9, pp.~7907--7914, Sept.~2024, 
doi: 10.1109/LRA.2024.3433203.

\bibitem{17}
R.~Bhirangi, M.~Gandhi, H.~Ha, Y.~She, and R.~R.~Salakhutdinov, 
``AnySkin: Plug-and-Play Modular Tactile Skins for Robots,'' 
\emph{arXiv preprint} arXiv:2311.01455, 2023.

\bibitem{18}
K.~Tomo, T.~Narumi, K.~Shimada, and Y.~Kawasaki, 
``uSkin: A Multi-Axis, High-Density, Compact Tactile Sensor for Robot Fingers,'' 
in \emph{Proc. IEEE/RSJ International Conference on Intelligent Robots and Systems (IROS)}, pp.~7608--7613, 2018.

\bibitem{19}
R. Li and E. H. Adelson, 
``Sensing and Recognizing Surface Textures Using a GelSight Sensor,'' 
in \emph{Proc. IEEE Conf. on Computer Vision and Pattern Recognition (CVPR)}, 
Portland, OR, USA, 2013, pp. 1241--1247, 
doi: 10.1109/CVPR.2013.164.

\bibitem{20} 
L. Sievers, J. Pitz, and B. Bäuml, 
``Learning purely tactile in-hand manipulation with a torque-controlled hand,'' 
in \emph{Proc. IEEE Int. Conf. Robotics and Automation (ICRA)}, Philadelphia, PA, USA, 2022, pp. 2745--2751.

\bibitem{21} 
H. Ichiwara, H. Ito, K. Yamamoto, H. Mori, and T. Ogata, 
``Contact-Rich Manipulation of a Flexible Object based on Deep Predictive Learning using Vision and Tactility,'' 
\emph{arXiv preprint arXiv:2112.06442}, 2022.

\bibitem{22}
K. Kawaharazuka, K. Okada, and M. Inaba, 
``Deep Predictive Model Learning With Parametric Bias: Handling Modeling Difficulties and Temporal Model Changes,'' 
\emph{IEEE Robotics \& Automation Magazine}, vol. 31, no. 4, pp. 81--99, 2024, doi: 10.1109/MRA.2022.3217744.

\bibitem{23}
R. Suzuki, H. Idei, Y. Yamashita, and T. Ogata, 
``Hierarchical Variational Recurrent Neural Network Modeling of Sensory Attenuation with Temporal Delay in Action-Outcome,'' 
in \emph{Proc. IEEE Int. Conf. Development and Learning (ICDL)}, Macau, China, 2023, pp. 244--249, doi: 10.1109/ICDL55364.2023.10364405.

\bibitem{24}
L.~Mack, F.~Grüninger, B.~A.~Richardson, R.~Lendway, K.~J.~Kuchenbecker, and J.~Stueckler, 
``Visuo-Tactile Object Pose Estimation for a Multi-Finger Robot Hand With Low-Resolution In-Hand Tactile Sensing,'' 
in \emph{Proc. IEEE Int. Conf. on Robotics and Automation (ICRA)}, Atlanta, GA, USA, 2025, pp.~12401--12407. 
doi: 10.1109/ICRA55743.2025.11127966.

\bibitem{25}
J.~Yin, H.~Qi, J.~Malik, J.~Pikul, M.~Yim, and T.~Hellebrekers, 
``Learning In-Hand Translation Using Tactile Skin with Shear and Normal Force Sensing,'' 
in \emph{Proc. IEEE Int. Conf. on Robotics and Automation (ICRA)}, Atlanta, GA, USA, 2025, pp.~5850--5856. 
doi: 10.1109/ICRA55743.2025.11127974.

\bibitem{26}
Z.-H.~Yin, C.~Wang, L.~Pineda, K.~Bodduluri, T.~Wu, P.~Abbeel, and M.~Mukadam,
``Geometric Retargeting: A Principled, Ultrafast Neural Hand Retargeting Algorithm,''
\emph{arXiv preprint} arXiv:2503.07541, 2025.
https://doi.org/10.48550/arXiv.2503.07541

\end{thebibliography}
\end{document}